\documentclass[sigconf]{acmart}

\usepackage{graphicx}
\usepackage{amsmath}

\usepackage{amssymb}
\usepackage{booktabs}
\usepackage{multirow}
\usepackage{bm}
\usepackage{arydshln}
\usepackage{bbding}
\usepackage{subcaption}

\AtBeginDocument{%
  \providecommand\BibTeX{{%
    \normalfont B\kern-0.5em{\scshape i\kern-0.25em b}\kern-0.8em\TeX}}}

\acmSubmissionID{3213}

\copyrightyear{2023} 
\acmYear{2023} 
\setcopyright{acmlicensed}\acmConference[MM '23]{Proceedings of the 31st
ACM International Conference on Multimedia}{October 29-November 3,
2023}{Ottawa, ON, Canada}
\acmBooktitle{Proceedings of the 31st ACM International Conference on
Multimedia (MM '23), October 29-November 3, 2023, Ottawa, ON, Canada}
\acmPrice{15.00}
\acmDOI{10.1145/3581783.3612439}
\acmISBN{979-8-4007-0108-5/23/10}

\begin{document}

\title{Scene-aware Human Pose Generation using Transformer}


\author{Jieteng Yao}
\affiliation{%
  \institution{Department of Computer Science and Engineering, Shanghai Jiao Tong University}
  \city{Shanghai}
  \country{China}
}
\email{yjt18122666691@sjtu.edu.cn}

\author{Junjie Chen}
\affiliation{%
  \institution{MoE Key Lab of Artificial Intelligence, Shanghai Jiao Tong University}
  \city{Shanghai}
  \country{China}
}
\email{chen.bys@sjtu.edu.cn}

\author{Li Niu}
\authornote{Corresponding authors.}
\affiliation{%
  \institution{MoE Key Lab of Artificial Intelligence, Shanghai Jiao Tong University}
  \city{Shanghai}
  \country{China}
}
\email{ustcnewly@sjtu.edu.cn}

\author{Bin Sheng}
\authornotemark[1]
\affiliation{%
  \institution{Department of Computer Science and Engineering, Shanghai Jiao Tong University}
  \city{Shanghai}
  \country{China}
}
\email{shengbin@sjtu.edu.cn}

\begin{abstract}
Affordance learning considers the interaction opportunities for an actor in the scene and thus has wide application in scene understanding and intelligent robotics. 
In this paper, we focus on contextual affordance learning, \emph{i.e.}, using affordance as context to generate a reasonable human pose in a scene. 
Existing scene-aware human pose generation methods could be divided into two categories depending on whether using pose templates. 
Our proposed method belongs to the template-based category, which benefits from the representative pose templates. 
Moreover, inspired by recent transformer-based methods, we associate each query embedding with a pose template, and use the interaction between query embeddings and scene feature map to effectively predict the scale and offsets for each pose template. 
In addition, we employ knowledge distillation to facilitate the offset learning given the predicted scale. 
Comprehensive experiments on Sitcom dataset demonstrate the effectiveness of our method.
\end{abstract}




\maketitle

\section{Introduction}
Affordance learning \cite{gibson1979ecological} is a long-standing and significant topic, which studies the possible interactions that an actor is allowed to have with the environment. 
In application, affordance learning covers numerous types of tasks, \emph{e.g.}, contextual affordance learning \cite{lopes2007affordance,ugur2011going}, functionality understanding \cite{zhu2015understanding, shiraki2014modeling}, affordance classification/detection/segmentation \cite{varadarajan2013parallel,grabner2011makes,eigen2015predicting}.
In this paper, we focus on contextual affordance learning, which uses affordance as context for the specified tasks, \emph{e.g.}, scene-aware human pose generation.
Learning such affordance not only allows actors to better interact in the scene, but also offers considerate feedback to the scene designers.
Therefore, contextual affordance learning has profound benefit for content analysis \cite{gupta20113d}, intelligent robotics \cite{lopes2005visual}, and scene understanding \cite{castellini2011using}.

\begin{figure}[t]
\begin{center}
\includegraphics[width=0.95\linewidth]{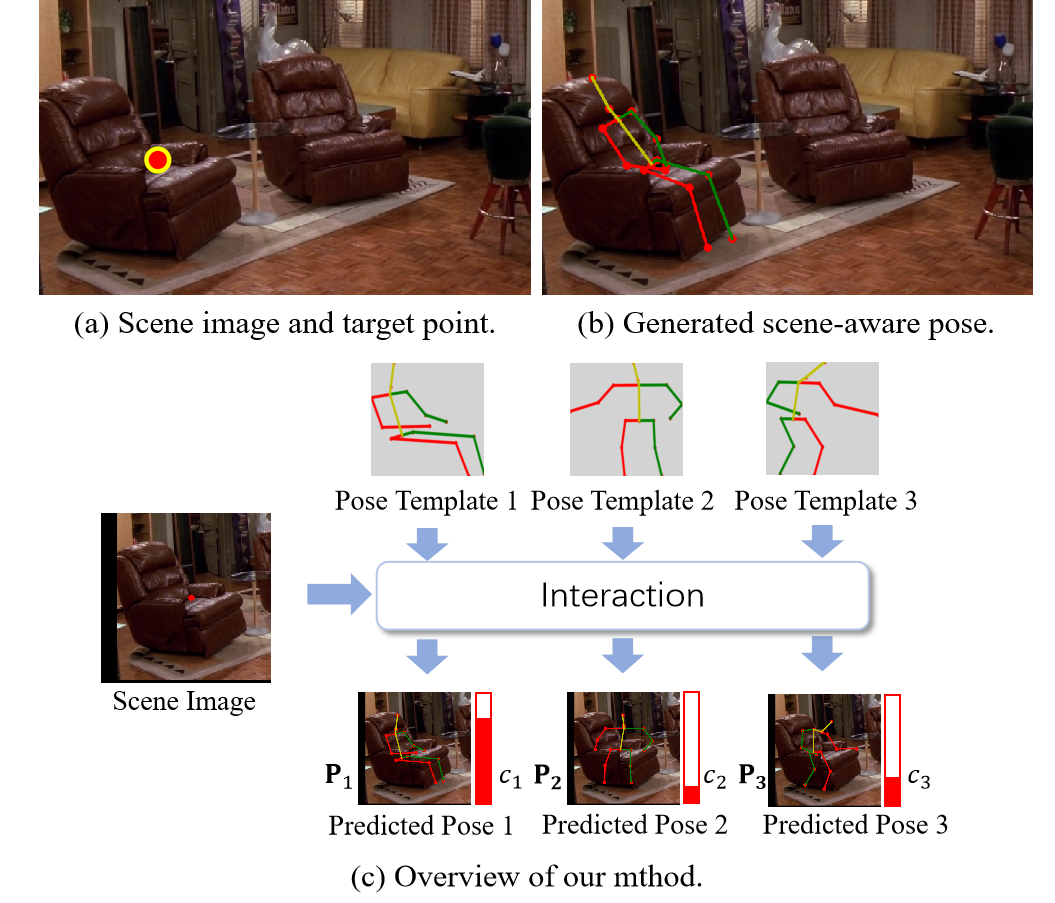}
\caption{
(a,b) Given a scene image and a target point, we aim to generate a reasonable human pose at this point. 
(c) Through the interaction between pose templates and scene image, our model predicts the pose $\textit{P}$ and corresponding compatibility score $c$ for each pose template. 
}
\label{fig:overview}
\end{center}
\end{figure}

In this work, we focus on learning the affordance between human and scene \cite{wang2017binge}, \emph{i.e.}, how we can put a human in a scene.
As shown in Figure~\ref{fig:overview} (a), given a 2D scene image and a target point, we need to generate a human pose that reasonably interacts with the scene at the target point.
Scene-aware human pose generation is actually a challenging task, because the model needs to understand the human, the scene, and the interaction between them.
For example, in the case of Figure~\ref{fig:overview} (a), the model should comprehend the function of sofa, know the possible poses of human body, place the human body on the sofa, and adjust the human keypoints according to the geometry of sofa reasonably.
As shown in Figure~\ref{fig:overview} (b), a hypothetical human could sit in the sofa with the back against the backrest and hams on the cushion.
Despite the significance of this task, there are only few efforts in this field. 
Previous methods and intuitive solutions of this task could be divided into two categories depending on whether using pose templates. Non-template-based methods are similar to pose estimation methods, which directly predict the location of each keypoint.
Template-base methods generate pose based on pre-defined templates. Wang \emph{et. al.} \cite{wang2017binge} constructed a dataset by collecting ground-truth poses at target points from video frames of sitcoms and proposed to use VAE \cite{kingma2013auto} to predict the scale and offsets of target compared with the chosen pose template. 

Inspired by the great success of transformer in a wide range of computer vision tasks \cite{dosovitskiy2020vit,carion2020end,cheng2021per}, we propose an end-to-end trainable network based on transformer.
Previous works \cite{carion2020end,cheng2021per} use query embedding to represent the potential object or category, while we use query embedding to represent the pose template.
As shown in Figure~\ref{fig:overview} (c), we set a number of pose templates in our model, which are prepared by using clustering algorithm to find representative poses in the training set.
Each pose template is associated with one template embedding to interact with the scene. Specifically, the $i$-th template embedding represents the $i$-th pose template.
Given a 2D scene image, we first extract its feature map via a backbone network and upsampling module, based on which the compatibility score of each pose template is predicted. Because there might be multiple pose templates which are reasonable for the input scene image, we introduce self-training strategy for compatibility score learning.
Then, each pose template interacts with the scene by feeding its query embedding and the scene feature map into two cascaded transformer modules \cite{vaswani2017attention} to obtain the pose scale and offsets. Specifically, the query embedding first interacts with the whole feature map to predict the scale according to the global context, and then interacts with the local feature map to predict the offsets according to the fine-grained context.
In this way, all plausible pose templates directly and parallelly interact with the scene image for predicting their scales and offsets.

For the model training, we use binary entropy loss to supervise compatibility score, and use MSE loss to supervise scale and offsets.
In order to facilitate the offset learning given the predicted scale, we introduce knowledge distillation in our model.
Moreover, we also supervise the refinement of templates that are not ground-truth but might fit the scene by adversarial loss.

We conduct comprehensive experiments on Sitcom affordance dataset \cite{wang2017binge} with in-depth analyses to demonstrate the effectiveness of our method.
Our contributions can be summarized as:
1) we propose an end-to-end trainable transformer network for scene-aware human pose generation;
2) we propose to facilitate the offset learning with knowledge distillation;
3) extensive experiments on Sitcom affordance dataset show the advantage of our method against state-of-the-art baselines.

\section{Related Works}
\subsection{Affordance Learning}
The concept of affordance \cite{gibson1979ecological} is first proposed by ecological psychologist James Gibson, which describes ``opportunities for interactions'' of environment.
Recently, affordance learning has included extensive applications, including contextual affordance learning \cite{lopes2007affordance,ugur2011going}, functionality understanding \cite{zhu2015understanding, shiraki2014modeling}, affordance classification \cite{varadarajan2013parallel,ugur2014bootstrapping}, affordance detection \cite{grabner2011makes,moldovan2014occluded}, affordance segmentation \cite{eigen2015predicting,roy2016multi}, etc.
Our work is more related to contextual affordance learning, which employs affordance relationships as context for associated tasks.
For examples, Lopes \emph{et al.} \cite{lopes2005visual} proposed to use contextual affordance as prior information to enhance gesture recognition and decrease ambiguities according to motor terms.
Castellini \emph{et al.} \cite{castellini2011using} employed affordance as visual features and motor features to benefit the object recognition task. 
Gupta \emph{et al.} \cite{gupta20113d} learned affordance in indoor images to detect the workspace according to human poses. 
Recently, Wang \emph{et al.}\cite{wang2017binge} adopted the scene affordance as context for human pose generation.
In this paper, we also focus on generating a scene-aware human pose as in \cite{wang2017binge}.

\subsection{Object Placement}
Human could be considered as a specific object, and thus our task is also related to the field of object placement, which is an important subtask of image composition \cite{niu2021making}.
Early methods \cite{tan2018and,tripathi2019learning,li2019putting,zhang2020learning,zhang2020and} mainly depended on explicit rules for placing foreground objects in background images, while recent methods \cite{liu2021opa,zhang2020learning,tripathi2019learning,zhou2022learning} employed deep learning for object placement.
To name a few, Lin \emph{et al.} \cite{lin2018st} proposed to use spatial transformer networks to learn geometric corrections to warp composite images to appropriate layouts.
PlaceNet \cite{zhang2020learning} predicted a diverse distribution for common sense locations to place the given foreground object on the background scene.
TERSE \cite{tripathi2019learning} proposed to train synthesizer and target networks in adversarial manner for plausible object placement.
Lee \emph{et al.} \cite{lee2018context} employed a two-stage pipeline to determine where to place objects and what classes to place.
Azadi \emph{et al.} \cite{azadi2020compositional} proposed a self-consistent composition-by-decomposition network to learn to place objects, based on the insight that a successful composite image could be decomposed back into individual objects. 
Most recently, Zhou \emph{et al.} \cite{zhou2022learning} treated object placement as a graph completion problem and proposed a graph completion module to solve the task.
The work \cite{fopa} proposed to identify the reasonable placements using a discriminative approach, by predicting the rationality scores of all scales and locations. 
In our task, the model needs to understand the scene and the human-scene interaction, and generate reasonable human poses from a large searching space, which is a very challenging task. 

\subsection{Human Pose Estimation and Generation}
Human pose estimation and generation has extensive applications and thus has attracted wide attention in decades.
As for human pose estimation, conventional methods \cite{dantone2013human, sapp2010cascaded, sun2012conditional, wang2013beyond, wang2008multiple} employed probabilistic graphical models or the pictorial structure models to represent the relations between human body parts.
Recent human pose estimation methods \cite{artacho2020unipose, cheng2020higherhrnet, lin2018learning, su2019multi, tang2019learning, li2021pose} have achieved significant progress by using deep learning models.
Technically, there are roughly two groups of pipelines for pose estimation: regression-based \cite{toshev2014deeppose,carreira2016human,sun2017compositional,luvizon2019human} and heatmap-based \cite{wei2016convolutional,chu2017multi,newell2016stacked,martinez2017simple}.
As for human pose generation, recent works generate human pose conditioned on past poses in video \cite{walker2017pose}, textual description \cite{hong2022avatarclip}, or video motion \cite{rempe2021humor}.
Except for \cite{wang2017binge}, there are few works generating human poses skeletons conditioned on 2D scene images.
\section{Method}

\begin{figure*}[t]
\begin{center}
\includegraphics[width=0.86\linewidth]{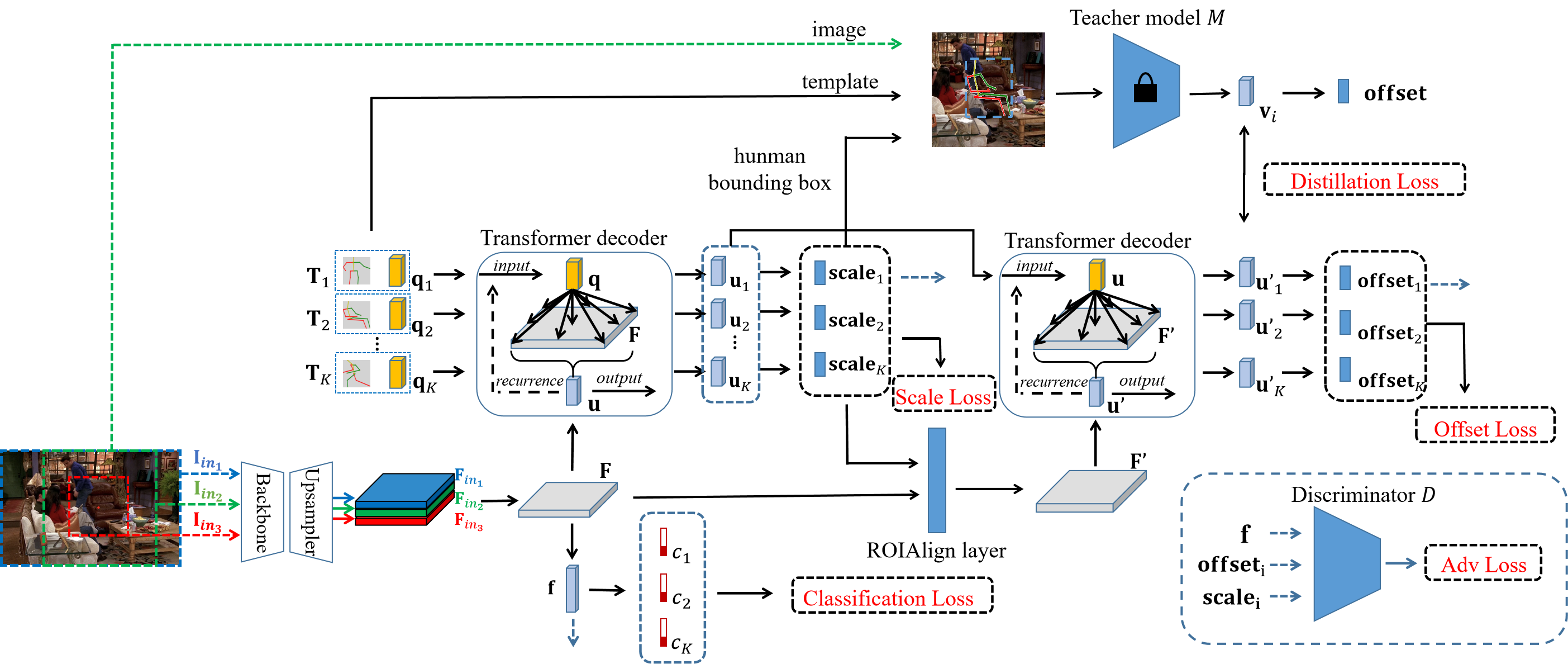}
\end{center}
\caption{Illustration of our method.
Our method consist of four modules: foundation network module, scale transformer module, offset transformer module, and distillation module.
The loss functions are marked in red.
More details can be found in Section \ref{sec:arch}.
}
\label{fig:method}
\end{figure*}

Given a full scene image $\textbf{I}_{full} \in \mathbb{R}^{3 \times H \times W}$ and a target point $\textbf{o} \in \mathbb{R}^2$, we need to generate a reasonable human pose with its enclosing rectangle centered at $\textbf{o}$.
As in \cite{wang2017binge,andriluka14cvpr}, the human pose $\textbf{P}$ is represented by $M=16$ keypoints (\emph{e.g.}, right ankle, pelvis), that is, $\textbf{P}=[\textbf{p}_1,\ldots,\textbf{p}_M]$ with $\textbf{p}_m \in \mathbb{R}^{2}$ being the $m$-th keypoint.
The whole pipeline of our method is shown in Figure~\ref{fig:method}, which uses $K$ pose templates and employs two cascaded transformer modules to model the interaction between each pose template and the input scene to produce the affordance pose. Each component of our method will be introduced in the following.

\subsection{Pose Template Construction} \label{sec:pose_template}
Considering that the definition of affordance learning in our task is how a hypothetical human could interact with the scene, we construct a number of representative pose templates in training set to facilitate the interaction modeling between the hypothetical human and scene image.
Specifically, we first normalize all ground-truth (GT) training poses. Then, we apply K-means clustering and further select $K$ representative cluster centers as $K$ pose templates $\{\textbf{T}_{i}\}_{i=1}^K$ by considering poses as $2M$-dim vectors and using Euclidean distance as distance metric ($K=14$ by default).

For pose normalization, we define the box centered at $(0,0)$ with side length $1$ as the unit box, denoted as $\textbf{B}_u$.
We also define a function $\textbf{B}_e(\cdot)$ to output the enclosing box of its input pose, \emph{i.e.}, by fetching the maximum and minimum x,y-coordinates of keypoints in the input pose.
For each GT training pose $\textbf{P}^*$, we compute its normalized pose by
\begin{equation}
    \mathcal{N}(\textbf{P}^*)=\mathcal{D}_{\textbf{B}_e(\textbf{P}^*) \rightarrow \textbf{B}_u}(\textbf{P}^*),
    \label{eq1}
\end{equation}
where $\mathcal{D}_{\textbf{B}_x \rightarrow \textbf{B}_y}(\cdot)$ is a linear deformation function defined by mapping box $\textbf{B}_x$ to box $\textbf{B}_y$ and could be used to deform the keypoints of input pose.
In this way, each normalized pose will have the unit box $\textbf{B}_u$ as its enclosing box, \emph{i.e.}, $\textbf{T}_i \in[-0.5,0.5]^{2M}$.
In our method, these pose templates will be refined to affordance pose according to the interaction with scene image.

\subsection{Scene Image Preparation}
When the scene image $\textbf{I}_{full}$ and target point $\textbf{o}$ are given, we prepare three forms of input images to leverage multi-scale information following \cite{wang2017binge}. As shown in Figure~\ref{fig:method},
the first input image $\textbf{I}_{in_1}$ is the whole image offering global context of scene image.
The second input image $\textbf{I}_{in_2}$ is a square patch centered at $\textbf{o}$ and the side length is the height of the scene image, which captures the scene context around the target location. 
The final input image $\textbf{I}_{in_3}$ is also a square patch but in half height to provide high-resolution information of scene context.
We re-scale three images to the same size $H_{in}\times W_{in}$, which is set as $H_{in}=W_{in}=224$ in implementation.
For each sample, the ground-truth label consists of three parts: class $i^*$, scale $\textbf{s}^*$, and normalized pose $\mathcal{N}(\textbf{P}^*)$. Class $i^*$ is the index of nearest pose template. We calculate the scale $\textbf{s}^*$ as the actual height and width divided by the actual height of $\textbf{I}_{full}$.

\subsection{Network Architecture}
\label{sec:arch}
As shown in Figure~\ref{fig:method}, our network consists of four modules:
1) a foundation network module, which generates feature map of the input scene image and predicts the compatibility score of each pose template;
2) a scale transformer module, which contains several transformer decoder layers as in \cite{carion2020end, cheng2021per} and an MLP prediction head;
3) an offset transformer module, which is similar to scale transformer module in architecture;
4) a knowledge distillation module, which is a pre-trained offset regression network.

\textbf{Foundation network module.}
Foundation network module consists of a backbone network, an upsampling network, and a classifier.
We employ ResNet-$18$ \cite{he2016deep} as our backbone network.
Because the spatial resolution of ResNet-$18$ output ($7\times 7$) is not enough for fine-grained context, we append it with an upsampling module to yield relatively high-resolution feature map.
Specifically, similar to the pixel decoder in MaskFormer \cite{cheng2021per}, we gradually upsample the feature map and sum it with the intermediate feature map of the corresponding resolution from the backbone network. 
As aforementioned, for each full scene image, we have three forms of input image $\textbf{I}_{in_1}$, $\textbf{I}_{in_2}$, and $\textbf{I}_{in_3}$, leading to three feature maps $\textbf{F}_{in_1}$, $\textbf{F}_{in_2}$, and $\textbf{F}_{in_3}$ after the foundation network module. 
Finally, three feature maps are concatenated and squeezed via $1\times 1$ convolutional layer to produce the scene feature map $\textbf{F}\in \mathbb{R}^{C \times H_f\times W_f}$, in which
$H_f=W_f=28$ in our implementation.
In this way, the scene feature map contains coarse context, fine context, and global context, which provide complementary information for later scene-aware human pose generation.
The feature map will pass a global pooling layer and then pass a full-connected layer to get the binary compatibility score for each of $K$ pose templates. 

\textbf{Scale transformer module.}
Given the feature map of scene image $\textbf{F}$ and each pose template $\textbf{T}_i$, we use transformer module to model the interaction between pose templates and scene, which can be formulated as
\begin{equation}
    \textbf{u}_i=\mathcal{T}_s(\textbf{q}_i,\textbf{F},\textbf{F}),
\end{equation}
where $\mathcal{T}_s(query, key, value)$ is a transformer module as in \cite{carion2020end, cheng2021per} and $\textbf{u}_i\in \mathbb{R}^d$ is the interacted embedding decoded from the $i$-th template embedding $\textbf{q}_i$.
Specifically, the transformer module consists of $3$ transformer decoder layers.
In the transformer module, the embedding of each pose template interacts with each pixel of scene feature map, which facilitates human pose scale prediction through the global context of the whole image.
We then use an MLP to predict the scale $\textbf{s}_i \in \mathbb{R}^{2}$ (two scale values along the x-axis and y-axis respectively) of each pose template based on the interacted embedding $\textbf{u}_i$.

\textbf{Offset transformer module.}
After getting the predicted scale, we compute the enclosing human bounding box and pay more attention to the features within the human bounding box. Specifically, we crop the feature map of the enclosing box region and resize it to the size of original feature map ($28\times28$) using a ROIAlign layer. The resized feature map is then used as the key and value of the transformer module, which can be formulated as
\begin{equation}
    \textbf{u}^{\prime}_i=\mathcal{T}_p(\textbf{u}_i,\textbf{F}^{\prime},\textbf{F}^{\prime}),
\end{equation}
where $\textbf{F}^{\prime}$ is the resized feature map within human bounding box.
The transformer module $\mathcal{T}_p$ consists of $1$ transformer decoder layer.
Then we use a MLP to predict the coordinate offsets $\bm{\Delta}_i \in \mathbb{R}^{2M}$.

\begin{figure*}[t]
\begin{center}
\includegraphics[width=0.8\linewidth]{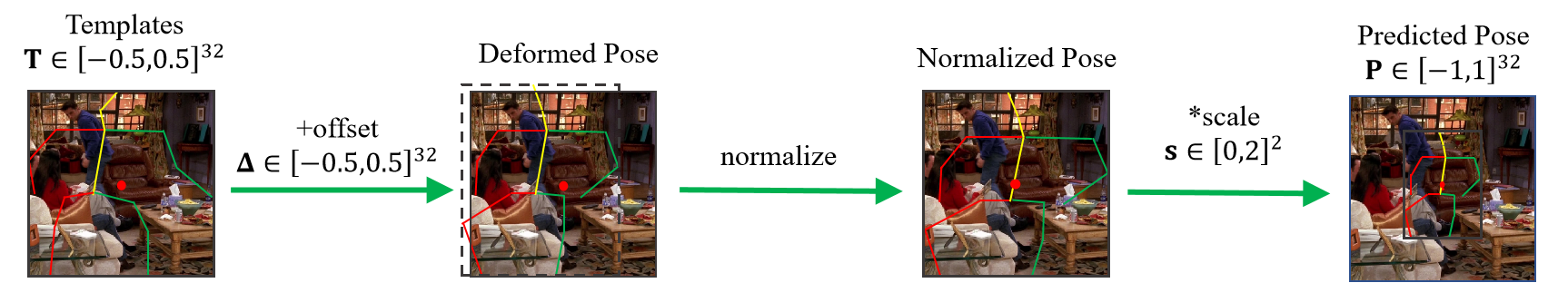}
\end{center}
\caption{
Illustration for the process of template refinement.
}
\label{fig:deform1}
\end{figure*}

Next we will show how to get the actual output pose prediction given a scene image.
As shown in Figure~\ref{fig:deform1}, the relative coordinate $-0.5$ (\emph{resp.}, $0.5$) on $x$-axis corresponds to the left (\emph{resp.}, right) boundary of input image $\textbf{I}_{in_2}$. Similarly, the relative coordinate $-0.5$ (\emph{resp.}, $0.5$) on $y$-axis corresponds to the bottom (\emph{resp.}, top) boundary of input image $\textbf{I}_{in_2}$.
We first add the offsets to the pose template (\emph{i.e.}, $\textbf{T}_i+\bm{\Delta}_i$) to obtain the raw affordance pose.
Then, we normalize this pose by $\mathcal{N}(\textbf{T}_i+\bm{\Delta}_i)$ (see Eq~ \ref{eq1}) to ensure that the center of predicted affordance pose is at the target point $\textbf{o}$.
Finally, we scale the pose with $\mathcal{N}(\textbf{T}_i+\bm{\Delta}_i)\cdot \textbf{S}_i$, where $\textbf{S}_i \in \mathbb{R}^{2M}$ denotes the scales of all $M$ keypoints and all $M$ keypoints share the same $x$,$y$-scale value $\mathbf{s}_i \in \mathbb{R}^{2}$. 
As for the range of offset prediction, we assert that any offset for normalized pose should not exceed a certain range, which is set to the size of normalized box, so we define the offset $\bm{\Delta}_i \in [-0.5,0.5]^{2M}$ for normalized pose. We also observe that the max height of ground-truth pose does not exceed twice the image height, so we set the scale $\textbf{s}_i \in [0,2]^{2}$ to ensure that the final predicted pose is in range of $[-1,1]^{2M}$.
Specifically, we refine each pose template to get the final pose by
\begin{align}
     \textbf{P}_i=\mathcal{D}_{\bm{\Delta}_i,\textbf{s}_i}(\textbf{T}_i)=\mathcal{N}(\textbf{T}_i+\bm{\Delta}_i)\cdot \textbf{S}_i.
 \end{align}

\textbf{Distillation module}
We introduce a distillation module to improve the performance of offset prediction. 
Because our model can be seen as a multi-task model, introducing a teacher model $M$ that focuses on predicting offsets given the scale and pose template can effectively help our model search for optimal solutions in a smaller range and converge faster. 
Specifically, we employ ResNet-$18$ \cite{he2016deep} as our teacher network.
During the pre-training phase, we first use ground-truth scale $\textbf{s}^*$ to derive the human bounding box, then we can fit the ground-truth pose template $\textbf{T}_{i^*}$ into the bounding box and get the heatmap of each keypoint. Finally, we concatenate the heatmaps and the scene image $\textbf{I}_{in_2}$ as the model input.
The teacher model predicts an offset vector, based on which we can get the corresponding predicted pose and use $L_2$ loss to supervise the model during pre-training.

During the training phase of our model, the keypoint heatmaps are generated by the ground-truth pose template $\textbf{T}_{i^*}$ and predicted scale from scale transformer module $\textbf{s}_{i^*}$. Provided with keypoint heatmaps, we use fixed pre-trained teacher model to extract the feature vector $\textbf{v}_{i^*}$. Recall that in our main network, offset transformer embedding $\textbf{u}^{\prime}_{i^*}$ accounts for offset prediction. 
Therefore, we realize knowledge distillation by narrowing the $L_2$ distance between offset transformer embedding $\textbf{u}^{\prime}_{i^*}$ and $\textbf{v}_{i^*}$.

\subsection{Training with Pose Mining}
In existing dataset Sitcom, each training sample is only annotated with one ground-truth pose, corresponding to one pose template. However, other pose templates might also be plausible for this scene image. 
Therefore, we employ self-training to mine more reasonable human poses in a scene. 
Specifically, we design a multi-stage training strategy. In the first stage, we assign each sample its nearest pose template as its ground-truth class and train our model. After that, we change the label to positive if the trained model of the previous stage predicts a compatibility score higher than a threshold (\emph{i.e.}, $0.7$). The training moves on to the next stage when the test classification accuracy begins to decrease, and stops when the labels no longer change. 
We apply binary cross-entropy loss to supervise the predicted compatibility scores:
\begin{align}
    \mathcal{L}_{\text{cls}}&=\frac{1}{K}\sum_i^K{w_i BCE(c_i, l_i)},
\end{align}
where $c_i$ is the compatibility score for the $i$-th pose template, $l_i$ is the label of the $i$-th pose template ($l_i=1$ if the $i$-th pose template is positive), $w_i$ is the weight to balance the quantity differences between positive and negative categories of the $i$-th pose template in dataset, and $BCE$ is the binary cross-entropy loss.

Though multiple pose templates could be classified as positive, the ground-truth poses and pose templates are unique for each sample. Therefore, we only apply the refinement loss to the ground-truth pose template.
The loss functions for the two refinement (scale and offsets) predictions could be formulated as
\begin{align}
    \mathcal{L}_{\text{offset}}&= {\| \mathcal{N}(\textbf{T}_{i^*}+\bm{\Delta}_{i^*})-\mathcal{N}(\textbf{P}^*)}\|^2,\\
    \mathcal{L}_{\text{scale}}&= {\| \textbf{s}_{i^*}-\textbf{s}^*}\|^2.
\end{align}

We apply the distillation loss to the output embedding of ground-truth pose template in offset transformer module, which can be formulated as
\begin{align}
    \mathcal{L}_{\text{dis}}&= {\| \textbf{u}^{\prime}_{i^*}-\textbf{v}_{i^*}}\|^2,
\end{align}
where $\textbf{u}^{\prime}_{i^*}$ is the output embedding of offset transformer decoder, $\textbf{v}_{i^*}$ is the output embedding of teacher model.

Besides the single ground-truth pose template, there could be other positive pose templates. We apply an adversarial loss to these templates to ensure the quality of output poses. We use $Q_{i}=[\mathcal{N}(\textbf{T}_{i}+\bm{\Delta}_{i}), \textbf{s}_{i}, \textbf{f})]$ to denote the concatenation of three vectors, 
in which $\textbf{f}$ is the feature vector after average pooling layer in foundation network module.
$Q_{i}$ is fed into the discriminator $D$ as input. The adversarial loss can be formulated as
\begin{align}
    \mathcal{L}_{\text{adv}}=&\mathbb{E}_{Q_{i^*}}[\log D(Q_{i^*}))]+\mathbb{E}_{Q_{\tilde{i}}}[\log (1-D(Q_{\tilde{i}}))],
\end{align}
where the discriminator $D$ is an MLP and predicts whether the input is from ground-truth template, $\tilde{i}$ is the index $i$ which satisfies $l_i=1$ and $i\neq i^*$.

Therefore, the total training objective of our model is
\begin{align}
    \min_{\theta_G} \max_{\theta_D}\quad \mathcal{L}_{\text{cls}}+\lambda_{\text{o}} \mathcal{L}_{\text{offset}} + \lambda_s  \mathcal{L}_{\text{scale}} + \lambda_{\text{adv}}  \mathcal{L}_{\text{adv}} + \lambda_{\text{dis}}
     \mathcal{L}_{\text{dis}},
    \label{eq:total_loss}
\end{align}
where $\theta_D$ includes the model parameters of discriminator $D$ and $\theta_G$ includes the other model parameters. $\lambda_{\text{o}}$, $\lambda_{\text{s}}$, $\lambda_{\text{adv}}$, and $\lambda_{\text{dis}}$ are hyper-parameters for balancing loss items. 
In our experiments, we set $\lambda_{\text{o}}=10$, $\lambda_{\text{s}}=10$, $\lambda_{\text{adv}}=10$, and $\lambda_{\text{dis}}=1$ via cross-validation. 

In the inference stage, we sort the predicted poses $\{(c_i, \bm{\Delta}_i, \textbf{s}_i)\}_i^{K}$ according to their compatibility scores $c_i$,
and select the poses with top $k$ compatibility scores as our final prediction.

\section{Experiments}
\subsection{Dataset}
We conduct experiments on Sitcom affordance dataset \cite{wang2017binge}. The training samples are collected from the TV series of “How I Met Your Mother”, “The Big Bang Theory”, “Two and A Half Man”, “Everyone Loves Raymond”, “Frasier” and “Seinfeld”, consisting of $25$k accurate poses over $10$k different scenes. The test samples are collected from “Friends”, consisting of $3$k accurate poses over $1$k different scenes. 
Each scene image is annotated with one or more target location points and each target point has one annotated ground-truth pose with $16$ keypoints.

\begin{figure*}[t]
    \centering
    \includegraphics[width=0.85\linewidth]{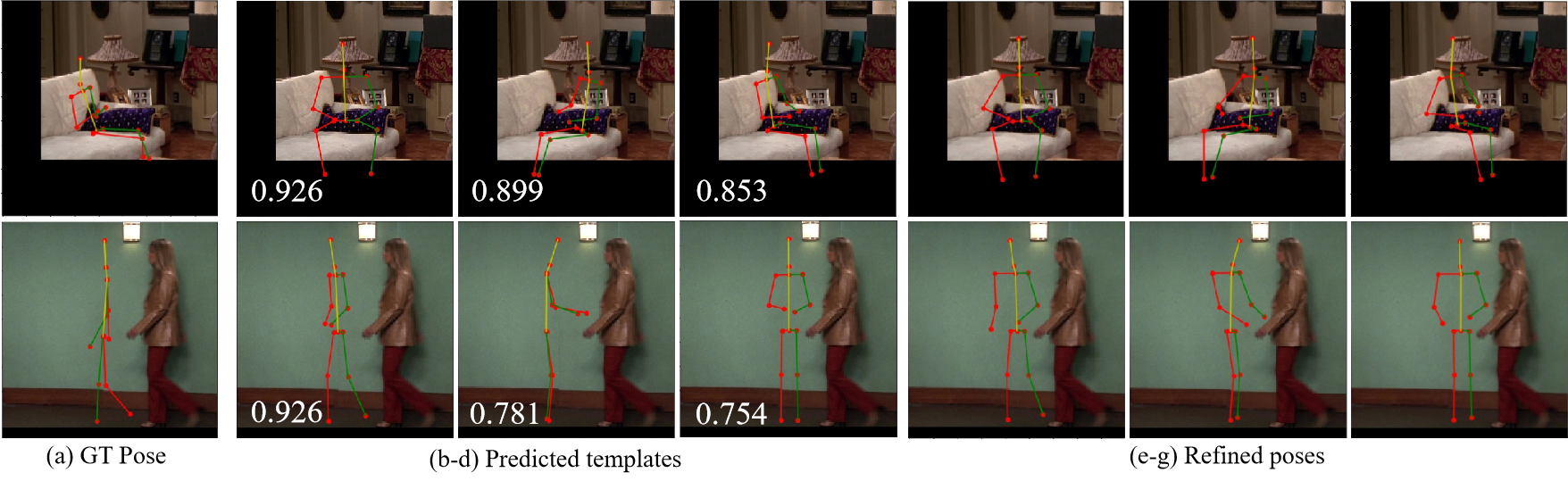}
    \caption{The in-depth visual analyses for our model prediction based on pose templates. 
    Column (a) shows the GT pose in the cropped patch of scene image $\textbf{I}_{2}$. Columns (b-d) show the top-3 predicted pose
templates with corresponding compatibility scores. Columns (e-g) show the predicted affordance poses refined from the top-3 pose templates.
    }
    \label{fig:viz_templates}
\end{figure*}

\subsection{Evaluation} \label{sec:eval}
We use three evaluation metrics in our experiments, \emph{i.e.}, PCK (Percentage of Correct Keypoints), MSE (mean-square error), and user study. 

PCK is a common metric used in pose estimation \cite{artacho2020unipose, li2021pose, cheng2020higherhrnet}, which indicates the percentage of correctly predicted keypoints that fall within a certain distance threshold of the ground-truth. In our task, we choose PCK@0.2 as in \cite{artacho2020unipose}, which refers to a threshold of $20\%$ of the torso diameter (the distance between left hip and right shoulder, corresponding to the $3$-rd point and the $12$-th point in Sitcom dataset).

MSE measures the mean-square error between the predicted pose and GT pose.
Specifically, we compute the mean-square error using the normalized coordinates which are divided by the heights of corresponding images, considering that the sizes of testing images may be quite different.

User study is conducted with $50$ voluntary participants by comparing the poses generated by different methods. 
For each test sample, every participant chooses the method producing the most reasonable pose. 
The score of each method is defined as the frequency that this method is chosen as the best one. 

\subsection{Implementation Details}
For the network architecture, we use ResNet-$18$ \cite{he2016deep} pretrained on ImageNet \cite{deng2009imagenet} as our backbone network and follow the setting of transformer module in \cite{cheng2021per} for the two transformer modules. 
For the MLP settings in the prediction head of two transformer modules, 
the first $2$ hidden layers in pose offset and scale prediction head have hidden dimension of $256$ with ReLU activate function. 
The last layer of offset prediction is followed by a tanh function, while the last layer of scale prediction is followed by a sigmoid function.
During the training stage, we apply SGD optimizer with learning rate $10^{-4}$ and weight decay $10^{-4}$. A learning rate multiplier of $0.1$ is applied to backbone network and $1.0$ is applied to other modules of model. The batch size is set to $8$. 
For the system environment, we use Python $3.7$ and Pytorch $1.8.0$ \cite{paszke2017automatic}. 
We conduct experiments on Ubuntu $18.04$ with Intel(R) Xeon(R) CPU $E5-2678$ v$3$ @ $2.50$GHz CPU and four NVIDIA GeForce GTX $1080$ Ti GPUs. 
The random seed is set as $0$ for all experiments unless stated otherwise.

\begin{table*}[ht]
    \centering
    \setlength{\tabcolsep}{5.5mm}{
    \begin{tabular}{c:c c c:c c c:c}
        \toprule
        \multirow{2}{*}{Model} & \multicolumn{3}{c:}{PCK $\uparrow$} & \multicolumn{3}{c:}{MSE $\downarrow$} & \multirow{2}{*}{User Study$\uparrow$} \\ 
        & Top-1 & Top-3 & Top-5 & Top-1 & Top-3 & Top-5 & \\\hline
        Heatmap &0.363 & - & - & 53.45 & - & - & 0.025\\
        Regression & 0.386 & - & - & 51.29 & - & - & 0.037\\
        Wang \emph{et al.} \cite{wang2017binge} & 0.401 & 0.432 & 0.458 & 46.65 & 44.49 &  43.17 & 0.225\\
        Unipose \cite{artacho2020unipose} &0.387&-&-&46.78&-&-&0.071\\
        PRTR \cite{li2021pose} &0.408&-&-&45.72&-&-&0.127\\
        PlaceNet \cite{zhang2020learning} & 0.060 & 0.082 & 0.107 & 377.78 & 375.92 & 371.60 & 0\\
        GracoNet \cite{zhou2022learning} & 0.380& 0.472& 0.473 & 49.60& 43.48& 43.35 & 0.110\\
        Ours &\textbf{0.414} & \textbf{0.498} & \textbf{0.533} & \textbf{44.86}& \textbf{41.91} & \textbf{39.42} & \textbf{0.405}\\
        \bottomrule
    \end{tabular}}
    \caption{The comparison of different methods on Sitcom affordance dataset. The best results are highlighted in boldface.}
    \label{tab:SOTA}
\end{table*}

\begin{figure*}[t]
\begin{center}
\includegraphics[width=0.88\linewidth]{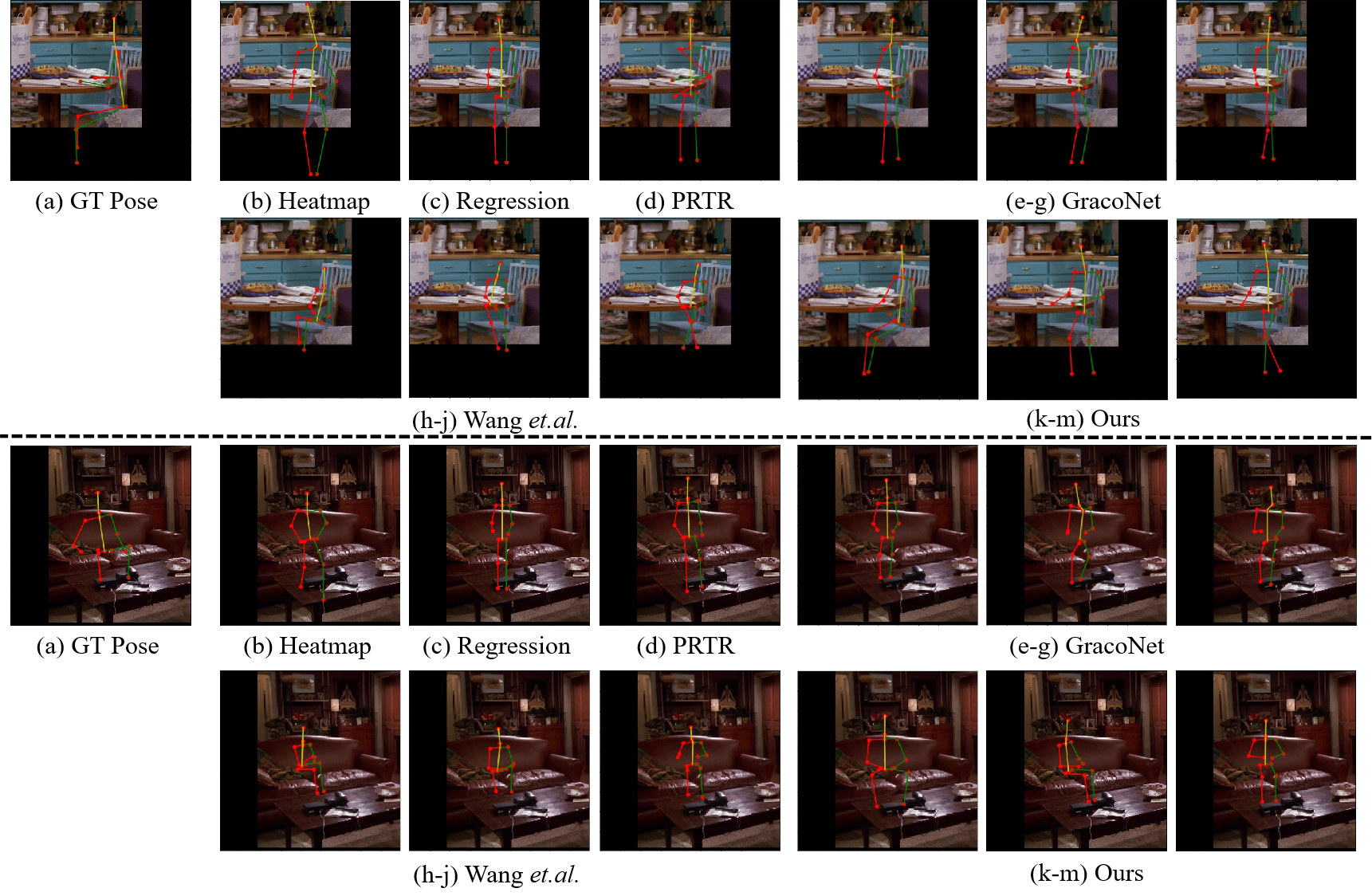}
\end{center}
\caption{Visual comparison of various methods on Sitcom affordance dataset.
In column (a), we plot the
GT pose in cropped scene image $\textbf{I}_{2}$
. In columns (b-m), we depict the affordance poses predicted by various methods.
}
\label{fig:viz_comp}
\end{figure*}

\subsection{Qualitative Analyses}
To provide in-depth analyses on how our model performs, we visualize the top-$3$ confident pose templates and the corresponding generated poses in Figure~\ref{fig:viz_templates}.
From the visualizations, we could see that our model could not only choose pose templates suitable for the input scene image but also refine them to reasonable affordance poses. 
Taking the first row as example, our model produces high compatibility scores for all \textit{sitting} pose templates, and then our model refines it to an affordance pose naturally \textit{sitting} on the sofa.
More qualitative analyses are left to Supplementary. 

\subsection{Comparison with Prior Works}
\textbf{Baseline Setting.}
We compare with the following representative baselines: two simple baselines, Wang \emph{et al.} \cite{wang2017binge}, pose estimation methods, and object placement methods.

1) Heatmap is a basic method producing keypoint heatmap.
Specifically, we concatenate three feature maps $\textbf{F}_{in_1}$, $\textbf{F}_{in_2}$, and $\textbf{F}_{in_3}$ after the ResNet-$18$ \cite{he2016deep} backbone network and upsampling module with upsampled resolution of $56\times 56$. Then, the concatenated feature maps are fed into a $1 \times 1$ convolution layer to produce an $M$-channel keypoint heatmap.

2) Regression is also a intuitive method directly predicting the coordinates of human pose joints.
We perform spatial average pooling on the concatenated feature maps which are the same as that in Heatmap baseline and send the pooled feature to a fully-connected layer to produce a $2M$-dim vector.

3) Wang \emph{et al.} \cite{wang2017binge} is a two-stage approach for scene-aware human pose generation based on pose templates.
In the first stage, it classifies the scene into one of $14$ pose templates which are the same as ours by sending the concatenated feature maps to a spatial average pooling layer and a linear classifier.
In the second stage, it employs VAE \cite{kingma2013auto} to produce the deformations based on latent code, one-hot pose label vector, and the concatenated feature maps.

4) Unipose \cite{artacho2020unipose} and PRTR \cite{li2021pose} belong to pose estimation methods. Unipose \cite{artacho2020unipose} is a unified framework based on "Waterfall" Atrous Spatial Pooling architecture.
PRTR \cite{li2021pose} is the first transformer-based model used in pose estimation tasks, which regards query embedding as keypoint hypothesis and generates the final pose by finding a match.

5) PlaceNet \cite{zhang2020learning} and GracoNet \cite{zhou2022learning} are two state-of-the-art methods in object placement task.
To adapt the model to our task, we remove the encoder of foreground in these models and replace foreground feature vector with a learnable embedding.

In these methods, we use the same backbone as ours to extract the image feature map.

\textbf{Quantitative Comparison.}
We first conduct quantitative comparison to precisely measure the discrepancies to GT poses in test set.
Specifically, in our method and Wang \emph{et al.} \cite{wang2017binge}, we obtain top-$k$ confident predicted poses with $k$ highest compatibility scores and calculate their PCK/MSE, after which the predicted pose achieving the optimal PCK/MSE is used for evaluation. 
in PlaceNet\cite{zhang2020learning} and GracoNet \cite{zhou2022learning}, there are only random vectors for generate multiple results, we sample $k$ times to get $k$ results and choose the predicted pose achieving the optimal PCK/MSE for evaluation.
Considering that the other baselines can not produce multiple confident predicted poses, we only report their top-$1$ metrics.
We summarize all the results in Table~\ref{tab:SOTA}.
Firstly, directly applying the simple methods (\emph{i.e.}, Heatmap and Regression) could only achieve barely satisfactory performances.
PlaceNet \cite{zhang2020learning} gets an inferior result because there is no reconstruction loss between generated pose and ground-truth pose during training.
The performances of GracoNet \cite{zhou2022learning} and Unipose \cite{artacho2020unipose} are hardly improved compared to Regression, indicating that these methods may need further improvement to fit this task.
 Wang \emph{et al.} \cite{wang2017binge} and PRTR \cite{li2021pose} could produce better results than baselines above, which proves the feasibility of using transformer and template-based method in this task respectively.
Finally, our method outperforms all the baseline methods especially in the top-$3$ and top-$5$ prediction. 

\textbf{Qualitative Comparison.}
Secondly, we conduct qualitative comparison to compare the visual rationality of various methods in different scenes. We present representative baselines, including Heatmap, Regression, PRTR \cite{li2021pose}, Wang \emph{et al.} \cite{wang2017binge}, and GracoNet \cite{zhou2022learning}. 
We show top-$3$ poses for the methods which can produce multiple results.
As shown in Figure~\ref{fig:viz_comp}, our method could produce overall better affordance poses in various scene images.
Specifically, Heatmap, Regression baselines, and GracoNet\cite{zhou2022learning} are prone to produce a standing human pose. One possible reason is that standing pose is the most common case, so they lack explicit knowledge about various human poses and tend to produce the average pose.
PRTR \cite{li2021pose} is better to some extent, but the predicted poses often lack integrity and rationality, probably focusing more attention on the location of single keypoint.
Wang \emph{et al.} \cite{wang2017binge} outperforms PRTR \cite{li2021pose} and GracoNet \cite{zhou2022learning}, producing more reasonable human poses due to representative pose templates.
Our model could further generate more diverse and plausible results. 
More qualitative comparisons are left to Supplementary.
\begin{table*}[t]
    \centering
    \begin{tabular}{c c c c c c c c c}
        \toprule
        \# templates & $K'$10 & $K'$20 & $K'$30 & $K'$40 & $K$20 & $K$14 & $K$10 & $K$7  \\
        \midrule
        Top-3 PCK$\uparrow$ &0.444	&0.466 &0.462	&0.447 & 0.476 & 0.498 & 0.493 & 0.479\\
        Top-5 PCK$\uparrow$ &0.477	&0.495	&0.505	&0.486 & 0.511 & 0.533 & 0.520 & 0.513\\
        Top-3 MSE$\downarrow$ & 48.37 &46.65	&46.78 & 50.92 & 44.14 & 41.91 & 42.32 & 42.84\\
        Top-5 MSE$\downarrow$ & 47.39 & 44.00 & 45.23 & 49.02 & 42.75 & 39.42 & 40.23 & 41.33\\
        \bottomrule
    \end{tabular}
    \caption{The performances of using different numbers of  pose templates. $K'$ denotes the number of templates from K-means, $M$ denotes the number of templates from selection.}
    \label{tab:varK}
\end{table*}

\begin{table*}[ht]
    \centering
    \begin{tabular}{c c c c c c c c}
        \toprule
        &classification & scale & offset & Top-3 PCK$\uparrow$ & Top-5 PCK$\uparrow$ & Top-3 MSE$\downarrow$ & Top-5 MSE$\downarrow$\\
        \midrule
        1&\textbf{u} &\textbf{u}&\textbf{u}&0.403 & 0.475 & 49.46&43.49\\
        2&\textbf{f} &\textbf{u}&\textbf{u}&0.461	& 0.493 &46.66&43.15 \\
        3&\textbf{f}+ST &\textbf{u}&\textbf{u}&0.475 & 0.496 & 44.54 & 43.07\\
        4&\textbf{f}+ST &\textbf{u}&$\textbf{u}^{\prime}$& 0.484 & 0.512 & 42.89&42.08\\
        5&\textbf{f}+ST &\textbf{u}&$\textbf{u}^{\prime}$+DS& 0.498 & 0.533 & 41.91&39.42\\
        \bottomrule
    \end{tabular}
    \caption{Ablation study on different modules. ST denotes self-training, DS denotes knowledge distillation.}
    \label{tab:ablation1}
\end{table*}
\begin{table*}[t]
    \centering
    \begin{tabular}{c c c c c : c c c c }
        \toprule
        &$\mathcal{L}_{\text{scale}}$ & $\mathcal{L}_{\text{offset}}$ & $\mathcal{L}_{\text{dis}}$ & $\mathcal{L}_{\text{adv}}$ & Top-3 PCK$\uparrow$ &Top-5 PCK$\uparrow$&Top-3 MSE$\downarrow$ & Top-5 MSE$\downarrow$ \\
        \midrule
        1&\Checkmark&&&&0.454&0.491&46.03&43.44\\
        2&\Checkmark&&&\Checkmark&0.456&0.508&45.93&43.05\\
        3&\Checkmark&\Checkmark&&&0.475&0.511&44.91&42.61\\
        4&\Checkmark&\Checkmark&&\Checkmark&0.484&0.512&42.89&42.08\\
        5&\Checkmark&\Checkmark&\Checkmark&&0.484&0.510&43.12&42.52\\
        6&\Checkmark&\Checkmark&\Checkmark&\Checkmark&0.498&0.533&41.91&39.42\\
        \bottomrule
    \end{tabular}
    \caption{Ablation study on loss functions.}
    \label{tab:ablation2}
\end{table*}

\subsection{Pose Template Analyses}
The pose templates play a pivotal role in our method. In order to investigate its impacts and characteristics, we conduct a series of experiments. 
We first group all annotated training poses into $10$, $20$, $30$, $40$ clusters using K-means algorithm respectively. 
Then, based on $30$ clusters, we further select $20$, $14$, $10$, $7$ representative clusters. We use $K'$ to denote the number of templates derived by K-means and $K$ to denote the number of templates derived by selection.
We summarize the results of various values of $K'$ in Table~\ref{tab:varK}, from which we could observe that as $K'$ increases, the performance first increases and then decreases. The cause of this may be that too few pose templates increases the refinement work of the model, while an excessive number of pose templates renders the classification process more challenging.
In manual selection cases, the performance first increases as $K$ decreases from $20$ to $14$, the decrease of $K$ after $14$ will lead to a decrease in model performance. It is probably because that the templates are redundant when $K$ is greater than $14$, while decreasing $K$ after $14$ will lead to the loss of semantic categories and the rest categories could not cover all data.
With an equivalent number of templates, manual selection yields better performance. We infer that it is probably because that the results from K-means algorithm lack semantic information, in which some pose templates share highly similar semantics.

\subsection{Ablation Study}
\textbf{Utility of Different Modules.}
As introduced in Section \ref{sec:arch}, we use two cascaded transformer modules to predict scale and offsets in sequence and predict the compatibility score based on global feature vector. 
Additionally, we incorporate knowledge distillation and self-training to facilitate learning.
To validate the effectiveness of these modules, we conduct the following experiments using different degraded versions of our model.
Firstly, we use a single transformer to perform all the classification, scale and offset prediction tasks (row 1). 
Then, we decouple the classification task and predict the compatibility score based on the feature vector (row 2).
We further employ self-training for the classification task (row 3).
Next, we introduce a new transformer with local feature map of human bounding box as input to predict offsets (row 4).
Finally, we add knowledge distillation to help predict the offsets (row 5).
The results are shown in Table~\ref{tab:ablation1}, from which we could observe that each change results in a performance improvement in the model.

\textbf{Utility of Different Loss Functions.} We conduct experiments on different loss functions. The results are shown in Table~\ref{tab:ablation2}. We observe that each loss item can improve the performance respectively. Generally, each loss makes up an important part, and they work together to guarantee the effectiveness of our method.

\section{Conclusion}
In this paper, we focus on contextual affordance learning, which requires to reasonably places a human pose in a scene.
Technically, we propose a transformer-based end-to-end trainable framework, which is based on pre-defined templates. We also introduce knowledge distillation to effectively help the offset learning.
Extensive experiments and in-depth analyses on Sitcom dataset have demonstrated the effectiveness of our proposed framework for scene-aware human pose generation.


\begin{acks}
The work was supported by the Shanghai Municipal Science and Technology Major/Key Project, China (Grant No. 2021SHZDZX0102, Grant No. 20511100300), and National Natural Science Foundation of China (Grant No. 62272298).
\end{acks}

\bibliographystyle{ACM-Reference-Format}
\bibliography{sample-base}


\end{document}


\title{Supplementary for Scene-aware Human Pose Generation using Transformer}

\author{Jieteng Yao}
\affiliation{%
  \institution{Department of Computer Science and Engineering, Shanghai Jiao Tong University}
  \city{Shanghai}
  \country{China}
}
\email{yjt18122666691@sjtu.edu.cn}

\author{Junjie Chen}
\affiliation{%
  \institution{MoE Key Lab of Artificial Intelligence, Shanghai Jiao Tong University}
  \city{Shanghai}
  \country{China}
}
\email{chen.bys@sjtu.edu.cn}

\author{Li Niu}
\authornote{Corresponding authors.}
\affiliation{%
  \institution{MoE Key Lab of Artificial Intelligence, Shanghai Jiao Tong University}
  \city{Shanghai}
  \country{China}
}
\email{ustcnewly@sjtu.edu.cn}

\author{Bin Sheng}
\authornotemark[1]
\affiliation{%
  \institution{Department of Computer Science and Engineering, Shanghai Jiao Tong University}
  \city{Shanghai}
  \country{China}
}
\email{shengbin@sjtu.edu.cn}



\maketitle

\appendix

In this supplementary, we will firstly present our used pose templates and more visualization results of our method in Section \ref{sec:analyses}.
We will show more qualitative comparison with baselines in Section \ref{sec:comparison}.
The complexity of our method will be reported in Section \ref{sec:complexity}.
We will provide more analyses for knowledge distillation module in Section \ref{sec:distill} and self-training strategy in Section \ref{sec:self-training}. 
Finally, we will discuss the limitation of our model in Section \ref{sec:limitation}.

\section{More Qualitative Analyses} \label{sec:analyses}
\begin{figure}
    \centering
    \includegraphics[width=0.95\linewidth]{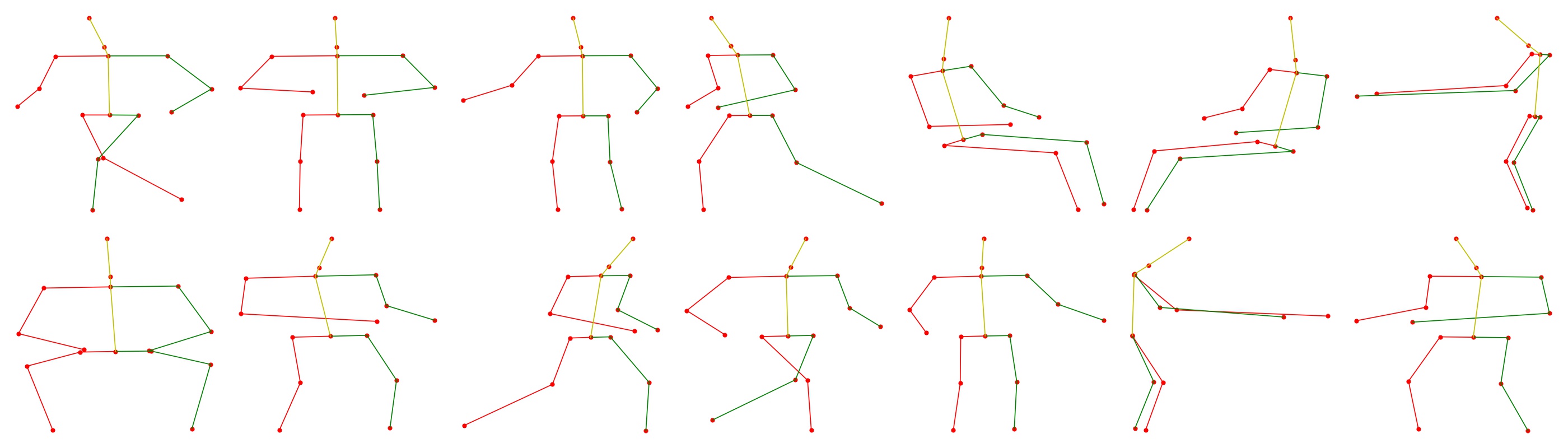}
    \caption{14 human poses templates in Sitcom affordance dataset used in our method.}
    \label{fig:templates}
\end{figure}

First, we present the pose templates we used in our model in Figure~\ref{fig:templates}. Recall that we first use K-means to cluster all training poses and then select $14$ representative poses from $30$ cluster centers as our final pose templates.  As shown in Figure~\ref{fig:templates}, each pose template has relatively independent semantics (\emph{e.g.}, sitting, running, standing), which facilitates our model learning.

\begin{figure*}[t]
\begin{center}
\includegraphics[width=0.95\linewidth]{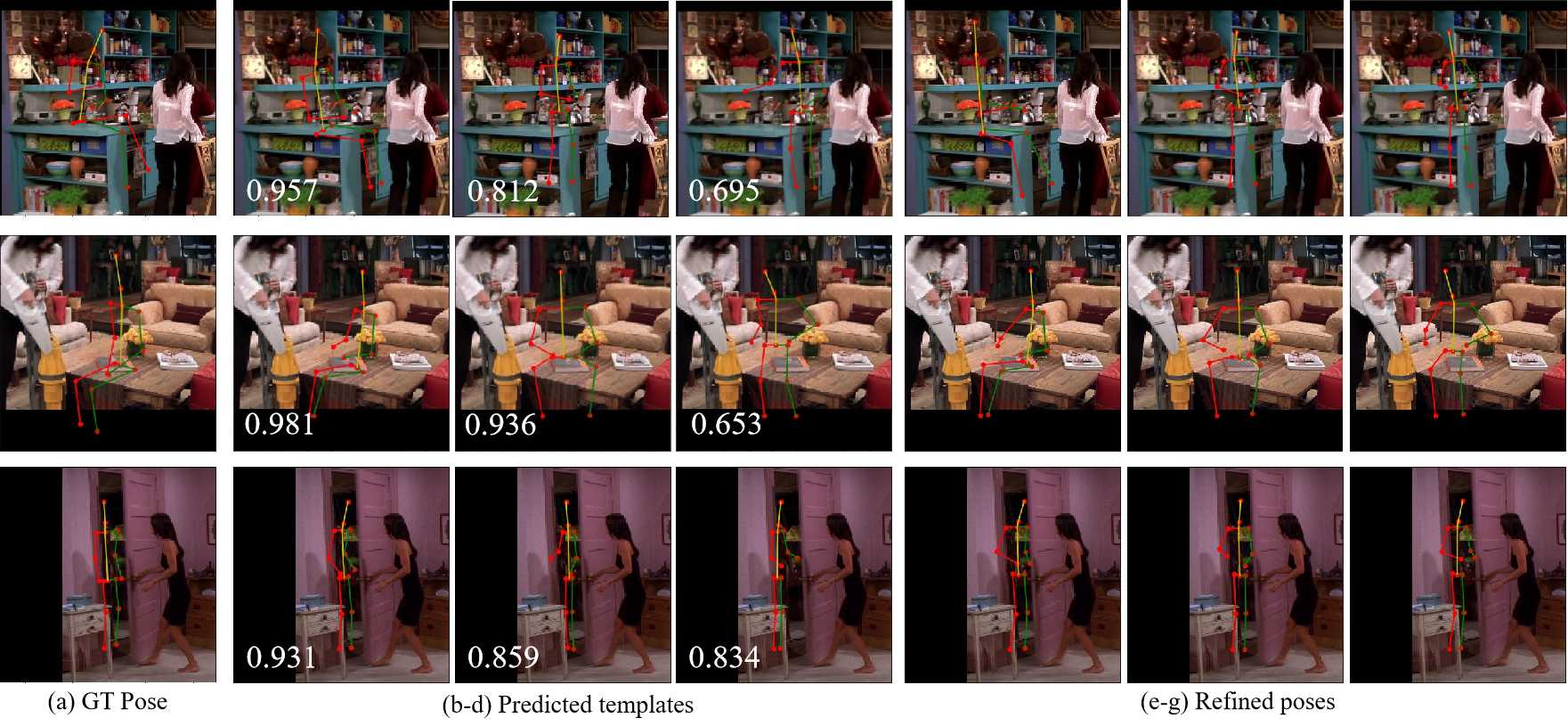}
\end{center}
\caption{Visual analyses for our model prediction based on pose templates. 
    Column (a) shows the GT pose in the cropped patch of scene image $\textbf{I}_{2}$. Columns (b-d) show the top-3 predicted pose
templates with the corresponding compatibility scores. Columns (e-g) show the predicted affordance poses refined from the top-3 pose templates.
}
\label{fig:supplement_top}
\end{figure*}

Next, we provide more visualization results of our method in Figure~\ref{fig:supplement_top}. 
From the visualizations, we can see that our method is able to identify the reasonable pose templates and refine them to plausible affordance poses. 
For example, in the second row, our model produces high compatibility scores for pose templates which are all relatively reasonable for the scene, \emph{i.e.}, ``sitting on the table and face others'', and further refines them to human poses naturally interacting with the scene.

\section{More Qualitative Comparison} \label{sec:comparison}
\begin{figure*}[t]
\begin{center}
\includegraphics[width=0.85\linewidth]{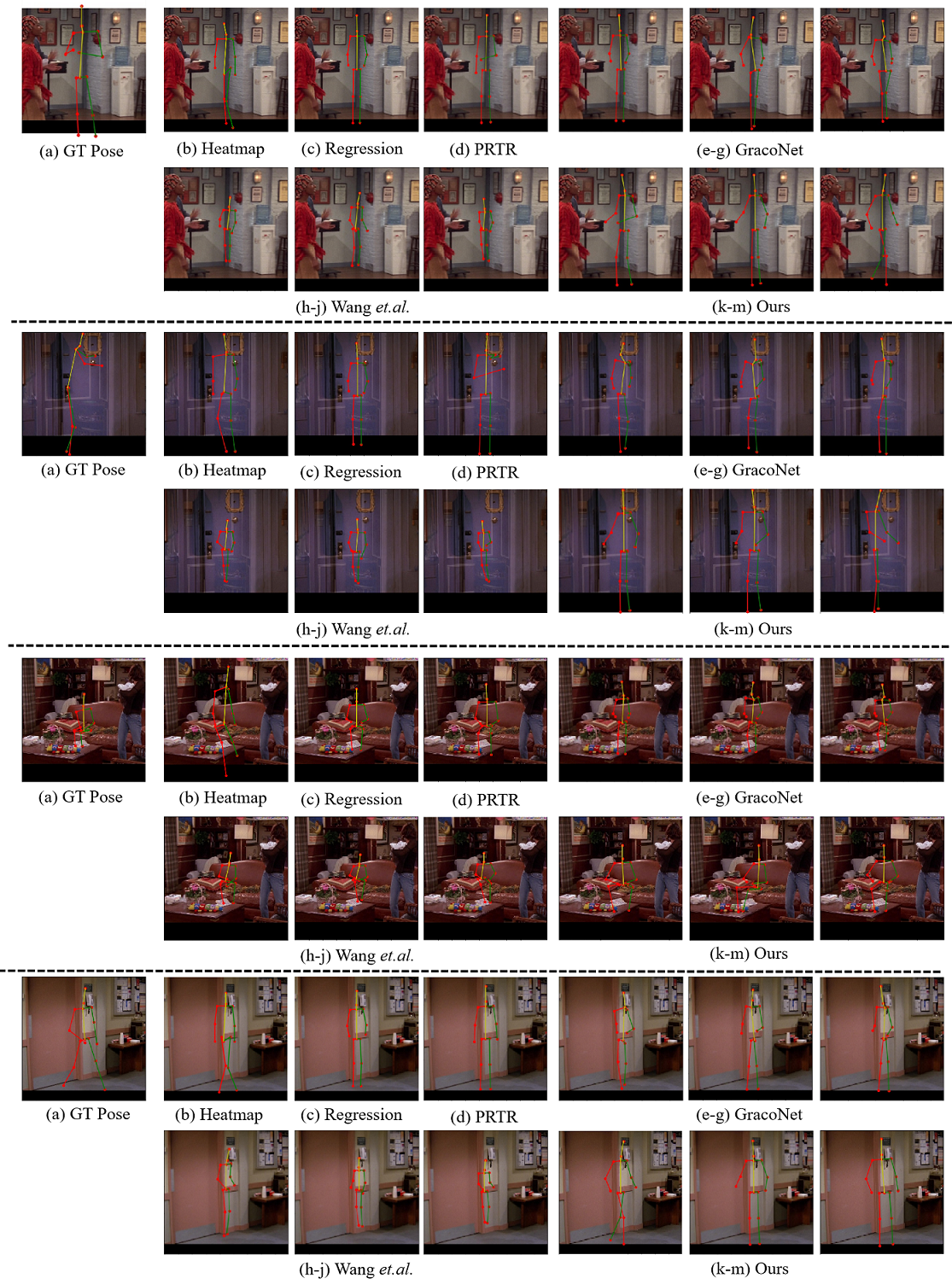}
\end{center}
\caption{Supplementary visual comparison of various methods on Sitcom affordance dataset.
The column (a) shows the GT poses in $\textbf{I}_{in_2}$.
The columns (b-m) show the affordance poses predicted by various methods.
}
\label{fig:supplement_comp}
\end{figure*}

In this section, we present more visualization comparison results in Figure~\ref{fig:supplement_comp}. 
As we can observe, our method can generate reasonable poses in different scenes, while other methods tend to generate unreasonable human poses.
For example, in the $1$-st sample, our method successfully generates ``interacting with others'' pose in the first two results. 
In the $2$-nd sample, our method generates ``reaching for the door handle'' poses in the first two results.
In the $4$-th sample, our method generates a ``walking'' pose, whereas other methods fail to generate this pose.

Here, we summarize the shortcomings of other baselines in generating poses. The poses generated by Heatmap method and GracoNet \cite{zhou2022learning} usually look unnatural. The poses generated by Regression method are prone to be the averaged ``standing'' pose. The poses generated by Wang \emph{et al.} \cite{wang2017binge} are often unreasonable in scale.
In contrast, our method could produce reasonable and plausible scene-aware human poses.

\section{Model Complexity} \label{sec:complexity}
We summarize the parameters count (\#Param), floating point operations (FLOPs), and inference time of different methods in Table~\ref{tab:complex}. The average inference time is calculated by running different methods on the entire test set with batch size of $1$ on a NVIDIA GeForce GTX $1080$ Ti GPU.
Concerning the parameters count, our model is smaller than GracoNet \cite{zhou2022learning} and similar to Wang \emph{et al.} \cite{wang2017binge} and PRTR \cite{li2021pose}.
Concerning the FLOPs, our model is smaller than Wang \emph{et al.} \cite{wang2017binge}.
Concerning the inference time, our model takes less time than PRTR \cite{li2021pose} and GracoNet \cite{zhou2022learning}, but takes a bit more time than Wang \emph{et al.} \cite{wang2017binge} though we use similar parameters count and FLOPs. One possible reason is that existing GPU computation has better efficiency optimization for fully-connected/convolutional layer against transformer decoder layer.
On the whole, our method has comparable complexity with the competitive baselines.

\begin{table}[t]
    \centering
    \begin{tabular}{c c c c}
        \toprule
        Model & \#Param & FLOPs & Inference Time \\
        \midrule
        Heatmap & 13.60M & 8.94G & 0.0141s\\
        Regression & 11.23M & 5.47G & 0.0067s\\
        Wang \emph{et al.} \cite{wang2017binge} & 24.54M & 10.94G & 0.0141s\\
        Unipose \cite{artacho2020unipose}& 19.62M & 5.55G & 0.0144s\\
        PRTR \cite{li2021pose}& 30.43M & 4.45G & 0.0341s\\
        PlaceNet \cite{zhang2020learning}& 12.27M & 1.83G & 0.0343s\\
        GracoNet \cite{zhou2022learning}& 46.57M & 1.91G & 0.0702s\\
        Ours & 29.23M & 9.40G & 0.0291s\\
        \bottomrule
    \end{tabular}
    \caption{The parameters count (\#Param), floating point operations (FLOPs), and inference time of various methods.}
    \label{tab:complex}
\end{table}

\begin{figure}
    \centering
    \includegraphics[width=0.95\linewidth]{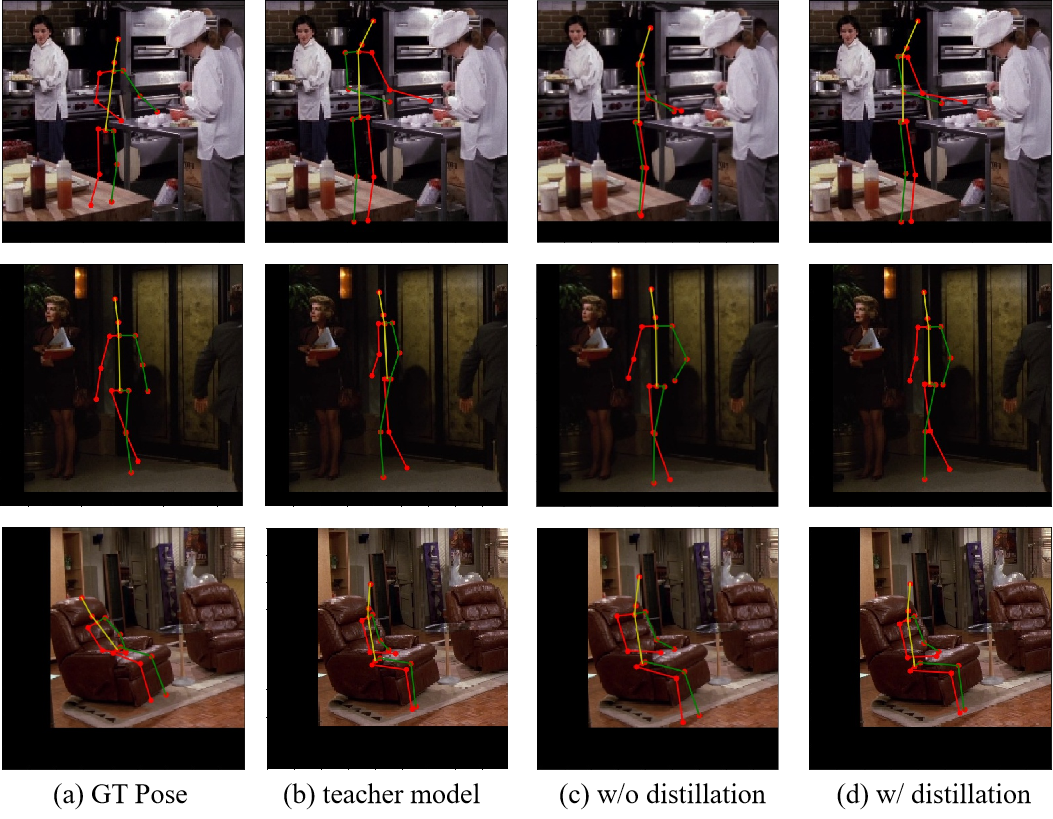}
    \caption{Visual comparison of teacher model and our model with/without distillation module.}
    \label{fig:dis}
\end{figure}

\section{Distillation Module Analyses} \label{sec:distill}
Recall that in our distillation module, we input the concatenated vector of image feature, pose template and scale into teacher model, then use the feature vector output by teacher model to supervise the offset transformer learning. 
To demonstrate the effectiveness of distillation module, we provide the predicted pose of teacher model, our method without distillation module, and our method with distillation module in Figure~\ref{fig:dis}. Specifically, we use the predicted image feature and scale of method with distillation module as the input of teacher model. All methods predict offsets based on the same pose template. It can be seen that our method with distillation module can predict poses more similar to the outputs of teacher model.
Taking the first row as an example, the pose predicted by our model with distillation module has a more similar angle to the output pose of teacher model, compared with our model without distillation module.

\section{Self-training Strategy} \label{sec:self-training}
\begin{figure}
    \centering
    \includegraphics[width=1\linewidth]{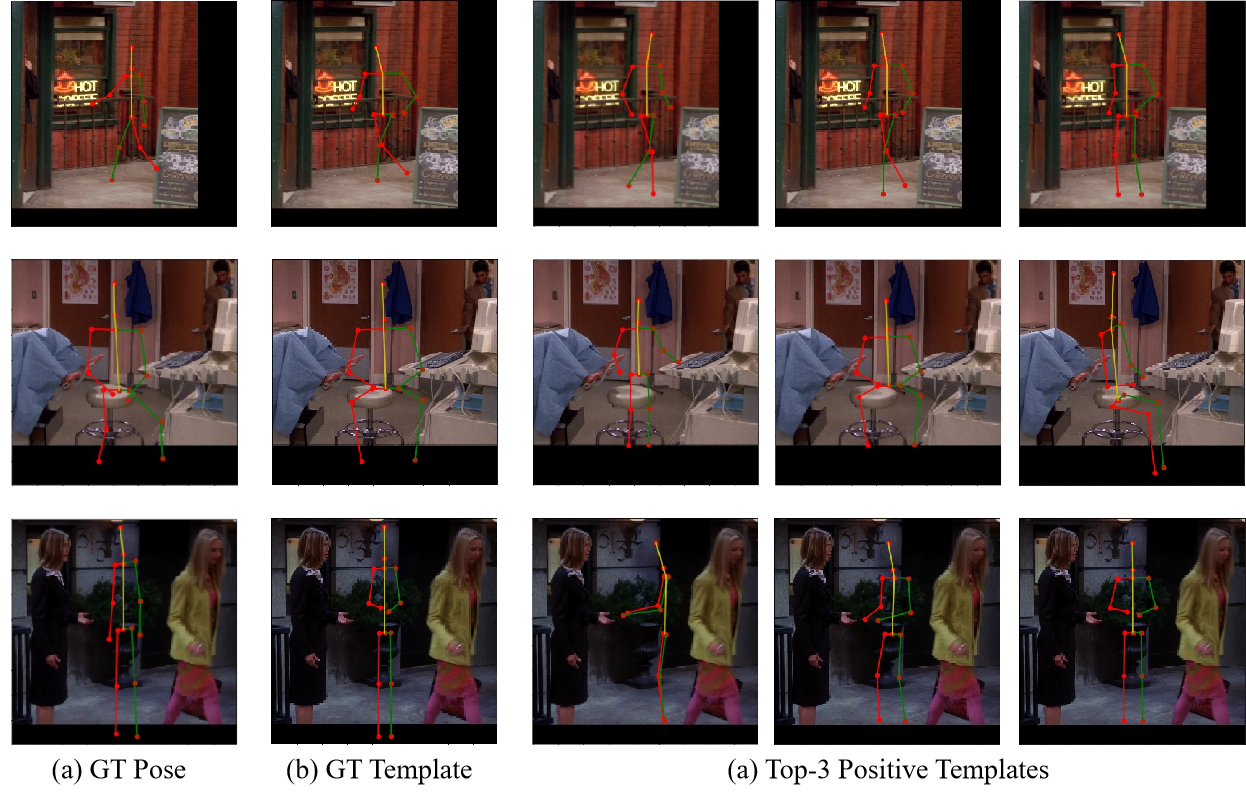}
    \caption{Visualization of ground-truth template and top-3 positive templates.}
    \label{fig:self}
\end{figure}

In our used dataset, given a scene image and a target point, there is only one annotated ground-truth pose templates. However, there could be multiple pose templates that are compatible with the scene image at this target point. Therefore, we introduce self-training strategy during model training.  
In this section, we study the impact of our self-training mechanism. 
We provide the original ground-truth templates for the scene image and the top-$3$ positive templates mined during self-training (the templates with compatibility scores higher than $0.7$ predicted by the trained model of the last training stage) in Figure~\ref{fig:self}. As we can see, the positive templates are also reasonable for the scene images and often include the ground-truth pose templates, which demonstrates that our self-training strategy can effectively mine the potential positive pose templates. 

\begin{figure}[t]
    \centering
    \includegraphics[width=1\linewidth]{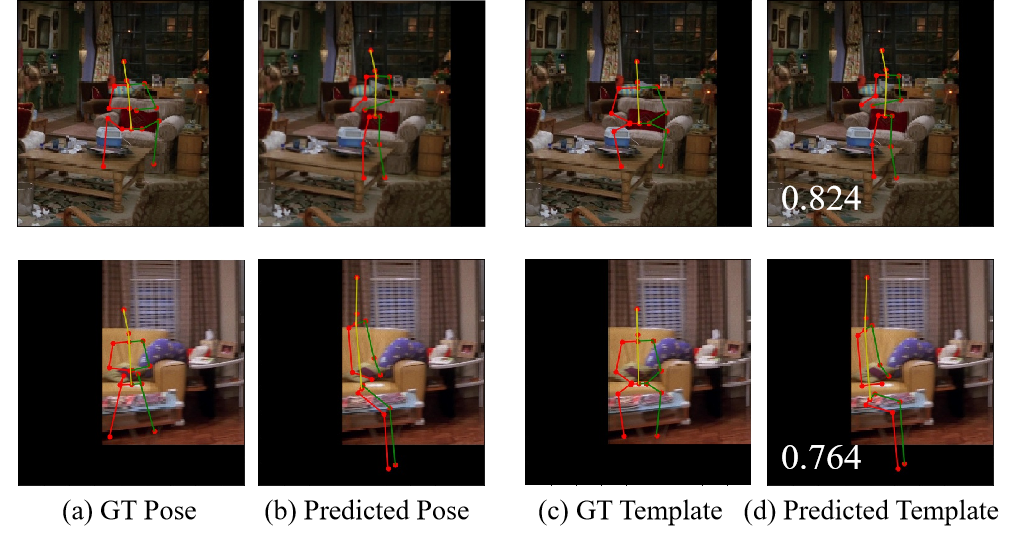}
    \caption{Illustration of our failure cases. 
    The column (a) shows the GT poses in the cropped patch of scene $\textbf{I}_{in_2}$.
    The columns (b) shows the predicted affordance poses.
    The columns (c) shows the GT pose templates.
    The columns (d) shows the predicted pose templates with the corresponding compatibility scores.}
    \label{fig:failure}
\end{figure}

\section{Limitation Discussion} \label{sec:limitation}
Our method is a template-based method, and thus the performance relies on two main aspects: compatibility score prediction and pose scale/offset prediction given a template.
As a consequence, our method is affected by the performances of these two aspects.
Although our method performs well in most cases, there still exist some failure cases. Specifically, our model may fail in choosing a suitable template which could enlarge the difficulty of pose refinement.
Besides, given a suitable and confident pose template, our model may fail to predict the proper scale or offsets.

One example of improper template selection is shown in the first row of Figure~\ref{fig:failure}. 
The model predicts a high compatibility score for an improper pose template (``standing''), so the refinement based on this template fails to produce a reasonable affordance pose.
One example of inaccurate prototype refinement is shown in the second row of Figure~\ref{fig:failure}. Although the model predicts the pose template correctly, it fails to predict the height scale of the pose. Therefore, the pose looks inconsistent with the scene context.

\bibliographystyle{ACM-Reference-Format}
\bibliography{sample-base}